\documentclass[11pt]{article}
\usepackage{acl}

\usepackage{times}
\usepackage{latexsym}
\usepackage[scaled=0.8]{beramono}

\usepackage[T1]{fontenc}

\usepackage[utf8]{inputenc}

\usepackage{microtype}

\usepackage{booktabs}
\usepackage{enumitem}
\usepackage{graphicx}
\usepackage{multirow}

\title{Educational Tools for Mapuzugun}

\author{Cristian Ahumada$^{1}$ \ \ \ Claudio Gutierrez$^{1}$ \ \ \ Antonios Anastasopoulos$^{2}$ \\
  $^{1}$Department of Computer Science, Universidad de Chile \\
  $^{2}$Computer Science Department, George Mason University \\
  \texttt{ahumada.860@gmail.com} \ \ \ \texttt{cgutierr@dcc.uchile.cl}  \ \ \ \texttt{antonis@gmu.edu} \\
}

\begin{document}
\maketitle
\begin{abstract}
Mapuzugun is the language of the Mapuche people. Due to political and historical reasons, its number of speakers has decreased and the language has been excluded from the educational system in Chile and Argentina.
For this reason, it is very important to support the revitalization of the Mapuzugun in all spaces and media of society. 
In this work we present a tool towards supporting educational activities of Mapuzugun, tailored to the characteristics of the language. 
The tool consists of three parts: design and development of an orthography detector and converter; a morphological analyzer; and an informal translator. We also present a case study with Mapuzugun students showing promising results.



\textbf{Short abstract in Mapuzugun:} Tüfachi küzaw pegelfi kiñe  zugun küzawpeyüm kelluaetew pu mapuzugun chillkatufe kimal kizu tañi zugun.

\end{abstract}

\section{Introduction}
Recent years have seen unprecedented progress for Natural Language Processing (NLP) on almost every NLP subtask. Along with research progress, several tools have been developed and are currently aiding millions of users every day. However, most of this progress is limited on a handful of languages~\cite{joshi-etal-2020-state}.
For example, learners of English can nowadays avail themselves to tools like Grammarly; English speakers can use Duolingo to start learning 38 languages, including Hawaiian, Navajo, as well as High Valyrian and Klingon.\footnote{As of March 2022.} The only option a Mapuzugun speaker would have in practice, though, would be to use language technologies in a language other than her own (likely Spanish).

Despite Duolingo's commendable inclusion of Hawaiian and Navajo for English speakers, and of Guaran\'i for Spanish speakers,\footnote{Which are due to immense efforts by the Indigenous communities themselves.} learning resources for Indigenous languages are hard to come by, let alone ones that incorporate language technologies in the educational setting in order to aid learners. In particular, it is undeniable that the development of NLP tools that reach the users lags further behind that NLP research itself~\cite{blasi-etal-21-inequalities}.

In this work, we develop a tool for educational use in an Indigenous language of south America, Mapuzugun. This tool was created by a speaker and instructor of the language 
and as such is tailored specifically to the instructional needs and linguistic characteristics of Mapuzugun.

Importantly, this work shows how linguistic research (grammars), minimal community resources (dictionaries), and NLP research (e.g. FST-based morphological analyzers) can be transformed into tools useful to Indigenous communities, in particular for efforts towards preservation and revitalization of endangered languages.
Our tool is publicly available through an online interface (in Mapuzugun and Spanish) at \url{crahumadao.pythonanywhere.com}.\footnote{Username: \texttt{epu} and
Password: \texttt{meli}} 

\section{The Mapuzugun Language}
Mapuzugun (iso 639-3: \texttt{arn}) is an indigenous language of the Americas spoken natively in Chile and Argentina, with an estimated 100 to 200 thousand speakers in Chile and 27 to 60 thousand speakers in Argentina \cite[41--3]{zuniga:2006}. 
It is an isolate language and is classified as threatened by Ethnologue, hence the critical importance of all documentary efforts.
Although the morphology of nouns is relatively simple, Mapudungun verb morphology is highly agglutinative and complex. Some analyses provide as many as~36 verb suffix slots~\cite{smeets1989mapuche}. A typical complex verb form may consist of five or six morphemes. See example in Table~\ref{tab:example}.

\begin{table}[t]
    \centering
    \small
    \begin{tabular}{l|p{5cm}}
    \toprule
        \textbf{Word} &  Kim mapuzuguyekümelleaiñ\\
        \textbf{Segmentation} & Kim mapu-zugu-yekü-me-lle-a-iñ\\
        \textbf{English Transl.} & We are indeed going to learn the Mapuche language.\\
    \bottomrule
    \end{tabular}
    \vspace{-1em}
    \caption{Segmentation of a Mapuzugun verb phrase.}
    \label{tab:example}
    \vspace{-1em}
\end{table}

Mapudungun has several interesting grammatical properties. It is a polysynthetic language in the sense of~\citet{baker:1996}; see \cite{loncon:2011} for explicit argumentation. 
As with other polysynthetic languages, Mapudungun has Noun Incorporation; however, it is unique insofar as the Noun appears to the right of the Verb, instead of to the left, as in most polysynthetic languages \cite{bakeretal:2005}.
One further distinction of Mapudungun is that, whereas other polysynthetic languages are characterized by a lack of infinitives, Mapudungun has infinitival verb forms; that is, while subordinate clauses in Mapudungun closely resemble possessed nominals and may occur with an analytic marker resembling possessor agreement, there is no agreement inflection on the verb itself.
One further remarkable property of Mapudungun is its inverse voice system of agreement, whereby the highest agreement is with the argument highest in an animacy hierarchy regardless of thematic role \cite{arnold:1996}.

Beyond morphology and other interesting typological properties, an additional challenge in the computational processing of Mapuzugun is the lack of a single standardized orthography. In particular, the community uses three different alphabets, namely the ``Unificado", ``Ragileo", and ``Az\"umchefe" alphabets.\footnote{See Figure~\ref{fig:alphabets} in Appendix~\ref{app:language}.}

\begin{figure*}[t]
    \centering
    \begin{tabular}{p{2.5cm}p{2.5cm}p{2.7cm}p{2.5cm}p{2.5cm}}
          \multicolumn{5}{c}{\includegraphics[scale=0.22]{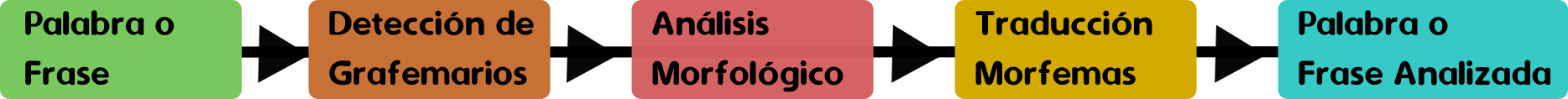}} \\
         \multirow{2}{*}{Input} & Orthography & Morphological & Morpheme & \multirow{2}{*}{Analysis} \\
         & Detector & Analysis & Translation & 
    \end{tabular}
    \caption{Pipeline of the full system.}
    \label{fig:pipeline}
\end{figure*}

\section{System Overview}
\begin{figure}[t]
    \centering
    \includegraphics[scale=.28]{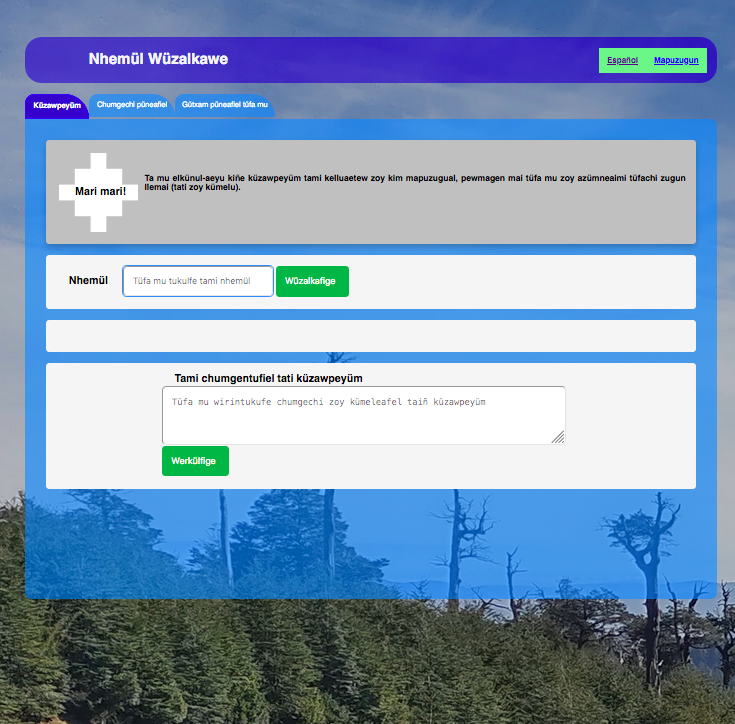}
    \caption{Screenshot of the
    user interface.}
    \label{fig:screenshot}
    \vspace{-1em}
\end{figure}

The system is comprised of the following components, with the pipeline shown in Figure~\ref{fig:pipeline}:
\begin{enumerate}[noitemsep]
    \item the orthography detector, which detects which of the three alphabets is used in the input;
    \item the orthography transliterator, which can convert between orthographies if conversion is needed;
    \item the morphological analyzer, which produces the possible segmentations of a word or phrase;
    \item the mapping of the analyzed morphemes to user-friendly notation/phrases; and
    \item the final presentation of the output.
\end{enumerate}
The user can use these tools through an interface available both in Mapuzugun and in Spanish. A screenshot of the landing page of the interface is shown in Figure~\ref{fig:screenshot}.

\section{Orthography Detection and Transliteration}

\begin{figure}[t]
    \centering
    \includegraphics[scale=.13]{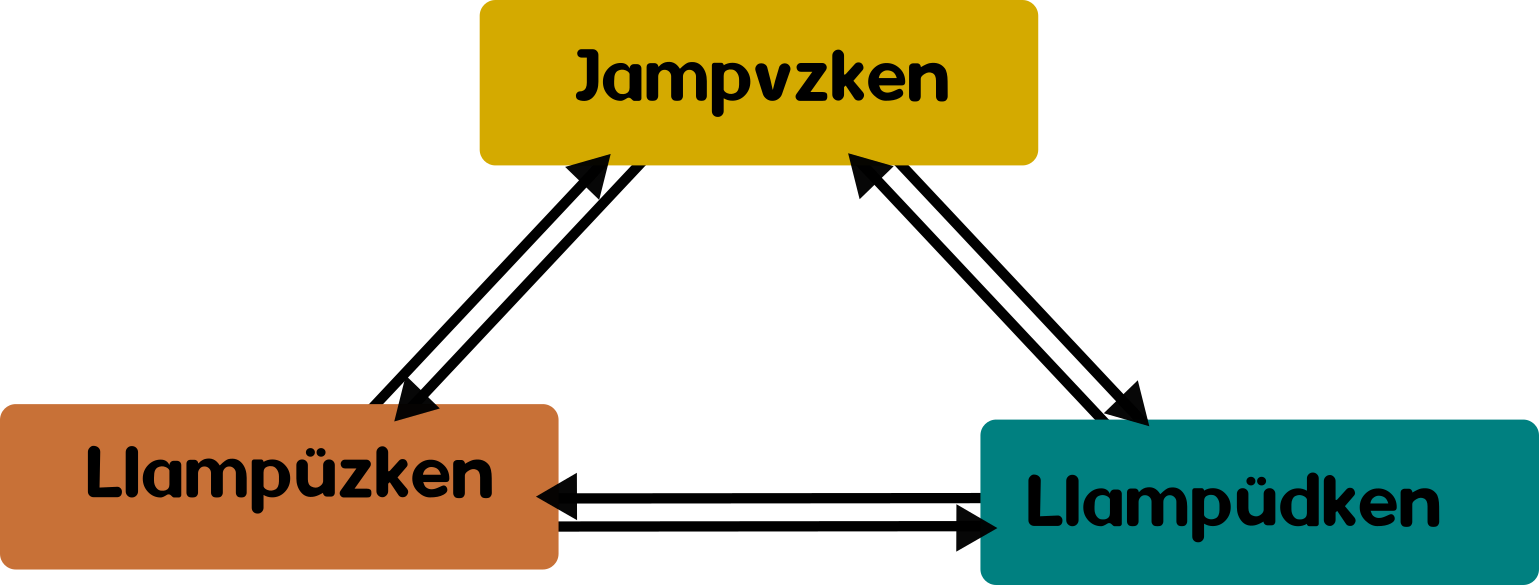}
    \caption{Conversions between orthographies for the Mapuzugun word for `butterfly'. Top: Ragileo; bottom left: Azümchefe; bottom right: Unificado.}
    \label{fig:conversions}
    \vspace{-1em}
\end{figure}

The differences between the three orthographies are showcased in Figure~\ref{fig:conversions}.
where “Jampvzken” is written in Ragileo, “Llampüdken” in Unificado and “Llampüzken” in Azümchefe, all three referring to the same Mapudungun phonetics of the English word ``butterfly''. This example shows the relationship between the `\texttt{J}' in Ragileo with the `\texttt{Ll}' in Azümchefe and Unificado.

We identified and constructed the conversion tables between these orthographies.
In total, for the Unified-Ragileo relationship, there are 10 differences that are shown in the Table~\ref{tab:uniragi}, in the following case Unified-Azümchefe there are 8 differences (Table~\ref{tab:uniazu}) and for the Ragileo-Azümchefe relationship there are 8 differences, outlined in Table~\ref{tab:ragiazu}.

Utilizing these conversion tables makes it straightforward to detect the orthography of any given input, by following a process of round-trip translation. For example, if we assume the input is in Ragileo, then if we convert to Az\"umchefe (or Unificado) and back to Ragileo and the final output is the same as the original input, then the input is declared to be Ragileo. If any of the intermediate translations fail it would have been exactly because our initial assumption of the input being in Ragileo was false. If no changes happen in the translation process, then 
all orthographies represent the input in a similar manner.

\paragraph{Orthography Converter}
Given the differences between the orthographies, special care must be taken in graphemes that have another grapheme as a substring. An example of this is the Unificado grapheme \texttt{Ng}, which also contains the grapheme \texttt{G}, which in turn is used in the same writing system for another phoneme. Or, cases in which three graphemes contain the same letter, such as the letter \texttt{"L"} in \texttt{L}, \texttt{Lh}, and \texttt{LL}. The only orthography that does not have this internal problem is Ragileo, because it uses unique letters for each Mapuzugun phoneme. This makes conversion from Ragileo to other orthographies straighforward, always taking care of the order of the transformations whose output can generate morpheme ambiguities during conversion.

The order that must be taken into account because if a morpheme is contained by another, it must first be disambiguated and then continue with other changes.
In the case of \texttt{Ng} and \texttt{g}, to go from Unificado to Ragileo or Azümchefe, as long as there is a \texttt{g} and there is no \texttt{N} preceding it, it can be changed to \texttt{Q}, therefore before making the transformations the \texttt{Ng} must be checked, saving the result \texttt{G}) in an auxiliary variable to be able to convert later all \texttt{G}s of the Unificado to \texttt{Q}. Once this last step is done, the auxiliary variable is removed and the \texttt{G} resulting from the change is put back.

\begin{figure*}[t]
    \centering
    \includegraphics[scale=0.53]{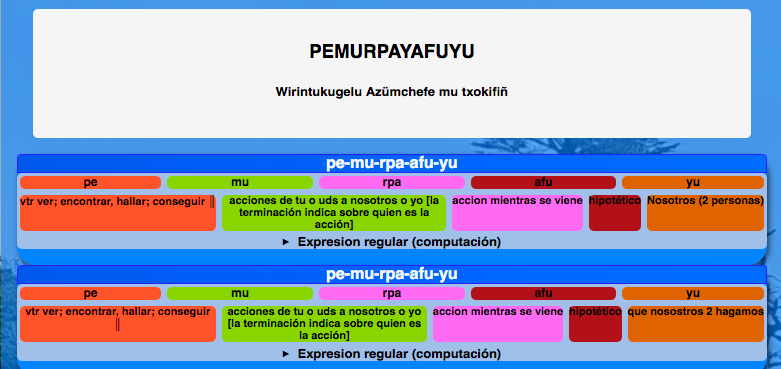}
    \caption{Segmentation of the word \texttt{"pemurpayafuyu"} as 
    presented by the tool.}
    \label{fig:word-example}
    \vspace{-1em}
\end{figure*}

\section{Morphological Analyzer}
The morphological analyzer is responsible for producing the possible segmentations: separating words into a composition of morphemes.

\subsection{Design}
The analyzer is implemented through series of regular expressions, based on established grammars of Mapuzugun~\cite{smeets1989mapuche,canumil,chiguailaf1972tagmemic}. As another source, the compilation that was made in \url{azümchefe.cl} of the grammar of the language~\cite{chiguailaf1972tagmemic} was taken.

We worked with hand-crafted sets of regular expressions that contain the morphemes of the language. These sets separate, by function: in verb root, noun/adverb/adjective, suffixes, and endings. In addition, the position plays an important role, because each of the morphemes has a particular slot~\cite{smeets1989mapuche}.

From these regular expressions, the chain of a word is traversed and possible derivations tree is generated. Only branches evaluated to be valid are passed on to the next “informal translator” step. The morphemes and their order must meet certain restrictions that have to do with the correct formulation of words in Mapuzugun, both in order, as mentioned before, but also in the compatibility of two morphemes being in the same word.

This module assumes input in the Ragileo orthography, therefore any word from another orthography must necessarily pass through the orthography converter. This decision has to do with Ragileo's advantage of 1-to-1 phoneme-to-grapheme mappings, making it easier to model morphemes.

\subsection{Informal Analysis Translator}


Once the segmentation is done, we implemented a module crucial
for deploying the tool in educational revitalization settings: the ``informal analysis translator''.
It assigns to each individual morphemes (or to combinations of them according to communicative role) a definition in plain Spanish.
The rationale was to simplify the definition as much as possible leaving out technical linguistic features and jargon. For the case of substantives, verbs and adjectives, the definition was taken from the Mapuzugun-Spanish dictionary \cite{perez:2015}.

As an example, we show the case of the word \texttt{txekayawkelai}. One of the possible segmentations is 
\textbf{txeka-yaw-ke-la-i}, with each component of the word being:

\begin{tabular}{r@{ }c@{ }l}
\textbf{txekan-}& : & vi \& vtr caminar, marchar, pasear  \\
   & & $||$ vtr medir con pasos \\
   & & \textit{to walk, to take a walk}\\
\textbf{-yaw-} & : & andar \textit{to go}\\
\textbf{-ke-} & : & habitualmente \textit{usually} \\
\textbf{-la-} & : & negación a modo "normal" \\
    && indicativo \\
    &&\textit{negation} \\
\textbf{-i} & : & el / ella \textit{he/she}
\end{tabular}
Given this,\footnote{"vi" and "vtr" correspond respectively to intransitive and transitive verb.} the goal is that the learner deduces
"el/ella no anda caminando habitualmente" \textit{``he/she does not usually go for walks"}.

The challenges of this informal analyzer are many.
Among them: how to 
give enough meaningful translations so that they can
match the initial experience of learners, but as well,
do not confuse them; how to deal with compositional morphemes (i.e. morphemes that have a different meaning when co-occurring than when occurring separately, for example transitions from second to first person); and how to include context to help the translation. We resolved these issues by relying on the expertise of an instructor of Mapuzugun.\footnote{One of the authors 
is a speaker and instructor of Mapuzugun.}

\section{Usability Studies with Learners}

The system (software) was tested on several groups of initial learners of Mapuzugun.

\begin{table*}[t]
    \centering
    \begin{tabular}{l|c|c|c|c|c|c}
    \toprule
  Word & Group 0 & Group 1 &  Group 2 & Group 3 & Group 4 & General \\ 
  \midrule
   \texttt{elukelafimu} & 2 / 2 &  2 / 1.6 & 2 / 1.86 & 2.75 / 2.86 & 2 / - & 2.33 / 2.14   \\
   \texttt{pemurpayafuyu} & 1 / 3 & 1 / 2 & 1.83 / 2.17 & 2.33 / 2.83 & 2 / - & 1.94 / 2.44 \\
   \texttt{kujinerkeei\~{n}mu} & 0 / 1.33 & 0.67 / 1.4 & 1.75 / 2 & 2.43 / 2.86 & 3 / - & 1.81 / 2.05 \\
   Phrase & 1 / 3 &  2 / 2.25 & 2.57 / 2.71 & 3 / 3 & 2.67 / - & 2.42 / 2.72 \\ \bottomrule
    \end{tabular}
    \caption{Summary of the study with learners.
    showing the mean performance of each group for each task word.
    Scale goes from 0 (wrong translation) to 
    3 (perfect translation). The pairs A / B mean: without / with the tool.}
    \label{tab:summary}
    \vspace{-1em}
\end{table*}

\begin{table*}[t]
    \centering
    \begin{tabular}{l|c|c|c|c|c|c}
    \toprule
   & Group 0 & Group 1 &  Group 2 & Group 3 & Group 4  & General \\ \midrule 
   Difficulty of use  & 2.67 & 2.71 & 1.86 & 1.86 & 3 & 2.33   \\
   Diff. of word transl. & 3.17 & 2.86 &   3.0 & 2.43 & 3 & 2.87 \\
   Diff. of phrase transl. & 3.33 & 3.71 & 2.71 & 3.14 & 1.67 & 3.0 \\
   Visual evaluation & 3.83 &  3.29 & 4.14 &  3.86 & 2.33 & 3.63 \\
   General evaluation & 3.17 & 4.0 & 4.71 & 4.29 & 3.33 & 4.0  \\
   \bottomrule
    \end{tabular}
    \caption{Summary of Usability Test. Scale 
    goes from 1 (low) to 5 (top).}
    \label{tab:usability}
    \vspace{-1em}
\end{table*}

\paragraph{Study Design}  

The first phase of the study design was to get access to study participants. As in the case of most endangered languages, it was difficult to identify test groups for various reasons.
First, most current Mapuzugun courses are informal, given different types of social organizations with a great variety of methodologies, contents, levels.
Second, students of Mapuzugun differ widely according to interests, degree of systematization and materials used. 
Third, there is a strong distrust by the interested community of learners in institutions, like academia, that historically have    ``used" aboriginal speakers as mere sources of information.

In a first preliminary round, more than 200 people (known to have been in courses or being students of Mapuzugun in the last 5 years) were contacted. From them, 30 people engaged to answer the questionnaire and from them, only 9 answers were obtained (3 of advanced knowledge of Mapuzugun).

With their feedback, the tool was refined. 
A second round was done by a public call in social networks related to Mapuzugun, and 32 people registered for the study, which were then classified in 5 groups: 

\begin{itemize}[nolistsep,noitemsep,leftmargin=*]
    \item Group 0. Beginners (6 people); 
    \item Group 1. Basic studies; able to greet but do not understand conversations (8 people); 
    \item Group 2. Studies: able to understand conversations (7 people); 
    \item Group 3. Studies; able to perform conversations (8 people); and 
    \item Group  4. Speakers from early infancy (3 people).
\end{itemize}

The experiment consisted of giving a small set of Mapuzugun words (and one phrase\footnote{The words are shown in Table~\ref{tab:summary}. The phrase was: \texttt{Pichikalu iñche , amukefun chillkatuwe ruka mew , fewla chillkatuwekelan.}}) to each participant. The task was to translate each word in Spanish, first without and then with the tool.

We additionally collected information on usability of the software tool: difficulty of use, difficulty to translate words, difficulty to translate phrases, evaluation of visual interface and finally, 
a general evaluation.
Last, we requested open-ended general qualitative feedback.

\paragraph{Translation Results}
Table~\ref{tab:summary} summarizes quality of the produced translations, with and without the tool, for each user group.\footnote{The translations were rated for accuracy by an instructor.}
For two words, \texttt{pemurpayafuyu} and
 \texttt{kujinerkeei\~{n}mu}, using the tool improves the translation capabilities for all user groups. The word  \texttt{elukelafimu} is a word that is typically accessible in basic levels of Mapuzugun, and hence, the segmentation plus the
 translation could have confused users (they
 realized that the word was more complex than
 they thought). Another encouraging sign is that the translation of the phrase also improved for the first three groups when using the tool. Last, we found that experienced
 learners (group 4), preferred not to use the tool because they
 felt secure in their knowledge.

\paragraph{Usability Results}
Table \ref{tab:usability} summarizes the scores received by the users (in a Likert scale). User groups 2 and 3 seem to be the ones showing less difficulty to use the tool, and also those that can take more advantage of it.  
Beginners got stuck with instructions (many were in Mapuzugun; they will also be provided in Spanish in future iterations) and ability to compose particles. We suspect that experienced speakers (group 4) probably did not invest effort because they did not need the tool.

All groups except experienced speakers rated the phrase as more difficult to translate than single words. The visual aspects of the interface and the tool in general mostly received very positive scores.



As a summary, our small study shows that, at its current stage of development, our tool is appropriate \textit{and} useful for intermediate learners. 

\paragraph{Qualitative Feedback}
We summarize here the qualitative feedback we received from user groups.

In general, all groups were particularly positive about the tool's presentation of the segmentation of the words. All groups were also very positive towards our informal translator that provides the explanations of each word segment.

In general, comments in the beginners' group (group 0) mentioned the difficulty to produce the translations, even though each part of the segmentation could be understood, a note that highlights the utility and importance of our proposed ``informal translation".
What was liked the most was the possibility of "see" in a graphical form the composition of words. This group also struggled with certain labeling words like VTR, VI, that are not widely known.

Users in group 1 positively mentioned the possibility to see the different segmentation options. Some people signaled that there should be examples of the usage.\footnote{Examples are provided as part of the documentation, but they probably did not find them.} 



Group 2 was the one that gave most comments. Some mentioned that a scenario when a morpheme occurs duplicated with different communicative functions was confusing. They also indicated that they would have wanted the ability to actually see the the correct translation, not just the segmentation and its explanation; unfortunately, the current state of MT for Mapuzugun does not allow this, but it provides a concrete avenue for future work. 

They also liked the segmentation and its explanation, and suggest that give the possibility to practice conjugation. 
On the other hand, words without context can be used in different forms and this could confuse beginners.

Last, there were comments about the choice of colors of the interface, as well as a suggestion for turning the tool into a mobile app.

Group 3 
suggested that beginners could get confused by the amount of options that are shown for certain words. Some of them mentioned that the program helped them to understand certain particles. They also mentioned the need of context for the words.
Regarding negative issues, some persons mentioned the need to have a translation besides morphemes, although one person liked the idea that you must make efforts to compose instead of receiving the translation immediately. 
Group 4 did not made relevant comments. 

It is worth noting, last, that many of the comments reflected the excitement that such a tool was even available for Mapuzugun.

\section{Related Work}
\paragraph{Computational Work on Mapuzugun}
Today there are various initiatives of computational linguistics on Mapuzugun. There is an orthographic normalizer and a morphological analyzer~\cite{chandia2012dungupeyem}, but its accuracy is low, since it is rule-based. Another aspect that could be improved is that, currently, there is no possibility of choosing the output alphabet, restricting it to only one form of writing. This is still inconvenient today, as there is still no agreement on orthography standardization. This implementation is based on a set of rules through regular expressions, with a finite state transducer, which have been released on the author's website. 

The purpose of another project, called AVENUE, in which the Universidad de la Frontera, the Intercultural Bilingual Education Program and the Language Technologies Institute of Carnegie Mellon University (CMU) collaborated, was to generate simple and low-cost translations, in addition to helping to preserve the Indo-American languages. This project first developed an alphabet that was used to transcribe (but not fully revise) a 170-hour audio corpus along with Spanish translations~\cite{duan-etal-2020-resource}, and last deployed prototype translation systems and base spell checkers that are available for OpenOffice. 

In the educational field, there is software to learn Mapuzugun called MAPU from a project at the Pontificia Universidad Católica de Valparaíso that also includes voice recognition to control the application, which works correctly, but is not robust to pronunciation~\cite{troncoso2012sistema}. This work also refers to another Mapuzugun-to-Spanish voice-text translation prototype, based on recordings, and to a chatbot from the Pandora project.


Last, we refer the reader to Appendix~\ref{app:sail} for an additional discussion of further computational work on other south American Indigenous languages.

\section{Conclusion and Future Work}
We have presented a system comprised of set of NLP tools appropriate for educational purposes in Mapuzugun, an Indigenous south American language, and we have demonstrated its usefulness through a small user study. Our study also provided a guide for future improvements. 
As more data will hopefully become available in Mapuzugun, we plan to incorporate more recent statistical machine learning components, both for the orthography converted and the morphological analyzer. We will also hopefully be able to deploy full-fledged machine translation systems to provide free-form translations of words or phrases to learners.
Many users would benefit by the incorporation of a text-to-speech component (as long as it is of high quality), that would also allow the teaching of Mapuzugun pronunciation.

Going further, the tool could be complemented with a system
that permits annotation of words and/or phrases in order to collect data for future tasks, as more users adopt it -- especially if language instructors use our tool in their courses. We are also hoping to create an offline version of the tool to make it accessible in areas with low connectivity. We will also attempt to incorporate any available corpora of Mapuzugun such as the those of~\citet{levin2000data} and~\citet{duan-etal-2020-resource} to use as educational examples.

We release our code\footnote{\url{https://github.com/crahumadao/kaxvkaam}} in the hopes that more Indigenous communities are able to use it to develop similar systems for their languages.

\section*{Ethical Considerations}

\looseness=-1 Working with endangered/Indigenous languages and language data, there is always substantial risk of unwittingly perpetuation of colonial harms~\cite{bird-2020-decolonising}. This is obviously an extremely complex issue, but according to \citet{bird-2020-decolonising} and other working in the space of NLP for endangered/Indigenous languages, perhaps the most critical aspect in working with Indigenous language data is that researchers actively develop meaningful relationships with members of these respective language communities. 

In our case, our work is lead by an instructor of Mapuzugun and member of the Chilean Mapuche community, who knows first-hand the oppression the Mapuche people have sufferred and the harms they have undergone by being forced to operate in Spanish. This work is also partially funded by a program dedicated to addressing the long-standing colonial harms in Chile, by specifically helping Indigenous students through their studies. 

We do not anticipate any serious harms by the development of our system, and we believe that the positive reception by the Mapuche volunteers who participated in our case study will be mirrored by its reception by the wider Mapuche community. It is also important, though, to acknowledge its limitations and make it clear that our tool is meant to be a companion tool for learning and can by no means substitute instructors of the language.

No indigenous language data were collected or are released through this project. We re-used existing, publicly available tools and corpora. The tool is provided for free: it is currently behind a simple username and password setting to ensure that its traffic is not overwhelming, so that the tool remains available to the Mapuzugun instructors and learners that need it the most (and who already have access to it).

\section*{Acknowledgements}
We are thankful to the Mapuzuguletuaiñ organization for their support and initial feedback, and to the volunteer Mapuzugun speakers who took part in our study.
The first author was funded by the Programa de Pueblos Ind\'{\i}genas, FCFM, Universidad de Chile. Antonios Anastasopoulos is generously supported by the NSF through project BCS-2109578.

\bibliographystyle{acl_natbib}
\bibliography{anthology,custom}

\clearpage
\appendix

\section{Notes on Mapuzugun}
\label{app:language}

In this section, to understand the context and the need for this work, it will be explained how Mapugugun went from being a language of a million speakers in the 16th century to becoming, according to UNESCO, an endangered language today. 


\paragraph{History of Mapuzugun}
The Mapuche people have their origin in the territory known as Wallmapu (which can be considered as the Mapuche Country \cite{millalen2006escucha}). This territory ranges from Coquimbo to Chiloé, also including areas on the "other side" of the mountain range, such as Neuquén, in a vast area demarcated by the Río Negro. Throughout this territory different denominations for this people can be found, shareing many cultural aspects \cite{millalen2006escucha}. This is the area in which the scope of the language known as Mapuzugun can be framed -- also in accordance with what the Spaniards defined at the end of the 16th century.

Mapuzugun, at its height, at the arrival of the Spaniards, was spoken by around a million people \cite{millalen2006escucha}. 
One of the first published books on this language is entitled "Art and Grammar of the General Language that runs throughout the Kingdom of Chile, with a Vocabulary and Confessionary" published in 1606 by Luis de Valdivia~\cite{alvarado2020glotopolitica}. In addition, toponyms with a clear Mapuzugun origin are still preserved, such as Huente Lauquen in the north, Puchuncaví, Curacaví, Pudahuel, Vitacura, with examples even in Puel Mapu (or what we know today as Argentina), and Chiloé in the south.

During the interaction of Mapuche with Spaniards during the Colony, the place of the Mapuzugun in all spheres of society can be appreciated, from the family, to international political relations, as were the Koyagtun (or Parliaments) mainly with the Spanish Empire, the Chilean and Argentine States, but also with the French, Dutch, and English. In all of these, the figure of the `\textit{lenguaraz}' stood out, who acted as a translator to try to faithfully reproduce the ideas that were held in Mapuzugun to foreign representatives.

It was during the construction of the Chilean and Argentine national states in the 19th century -which initially did not include Mapuche territory- with the so-called "Campaign of the Desert" and "Pacification of Araucanía", when these states politically subjected the Mapuche people. Along with this, as part of the construction of the identities of Chile and Argentina, space was taken away from Mapuzugun and the Mapuche culture through schools and the church, which, through evangelization and punishment, denied indigenous identity along with their language.

Then at the beginning of the 20th century, after that territorial dispossession, there was a strong Mapuche migration to the most important cities of Chile, in search of better living conditions. This translated into cultural loss, often due to racism and discrimination. However, some efforts were made by the Mapuche themselves to maintain the culture and language, as shown by publications such as those by~\citet{cona2019} and~\citet{manquilef1911comentarios,manquilef1914comentarios}, which were written in both Mapuzugun and Spanish.

During the dictatorship and since the 90's, the Mapuche people began to have a greater political position. With this, the Mapuche language was recovered hand in hand with a recovery of identity in various areas, in addition to maintaining territories in which Mapuzugun is spoken as the first language. Today, according to the 2017 census, the majority of the Mapuche population would be in Santiago, but most do not speak or understand their language.

Today, there are various organizations that offer courses or tools that contribute to the revitalization of Mapuzugun. These instances have a milestone in a march that is organized during February, within the framework of the commemoration of the "International Mother Language Day"~\cite{uatvcl}, having, as a movement, important demands such as the officialization of Mapuzugun~\cite{lengua2016gomez}.

Various sources estimate the number of Mapuzugun speakers to be between 100,000 and 300,000~\cite{encuesta}.\footnote{\url{ https://news.un.org/es/story/2019/04/1454571}} They constitute about~5 to~10\% of the Mapuche population (1,745,147), who make up 9.9\% of the total population of Chile (17,574,003).

According to UNESCO, a language is in danger when it is no longer used, when it is used in fewer areas and when it is no longer transmitted. From this it is stated that "about 90\% of all languages ``could be replaced by the dominant ones by the end of the 21st century". All this, added to insufficient documentation, generates that there are extinct or endangered languages. , which are unrecoverable~\cite{aronoff2020handbook}.

There are six degrees to define the state of danger of these languages. Within this classification is the Mapuzugun, referred to as Huilliche and, both in Chile and Argentina, as Mapuche, as can be seen in the Unesco Atlas~\cite{moseley2012unesco}. They are in grade 1 (“In a critical situation”, Huilliche), 2 (“Seriously in danger or threatened”, Mapuche, Argentina) and 3 (“Clearly in danger or threatened”, Mapuche, Chile).

But not only through UNESCO, research has been carried out on the state of the Mapuzugun. There are also various studies from the area of sociolinguistics to understand certain current language processes and their incorporation into public policies~\cite{lengua2016gomez,loncon2010derechos,catrileo2017diccionario,wittig2009desplazamiento,lagos2012mapudungun,olate2013interactividad} [46].

\paragraph{Typological Notes}
Linguistically, Mapuzugun is defined as an agglutinative and polysynthetic language, which means that its expressions have a main root to which defined and distinguishable suffixes are added to form phrases. For example the word Kim mapuzuguyekümelleaiñ, which is explained in Table 2.1. Examples such as English, Chinese or Spanish are not in this category, and therefore the processing techniques used in those languages differ from the techniques that could be used for Mapuzugun.

Before colonization, Mapuzugun is described as a purely oral language. Today, until recently it was not formally taught or used by public and educational institutions in Chile. This has meant that it does not have a standardization of its writing or spelling. Today there are different ways of writing it and also, territorial orthographic variations, because in different regions there are phonetic differences for certain sounds and that translates, in general, into different writings. Today there are three main graphemaries to write Mapuzugun: Ragileo, created by Anselmo Raguileo in 1985; Unified, created by María Catrileo in 1989; and Azümchefe, created by Necul Painemal for CONADI (National Corporation for Indigenous Development) in 2008.

Among these graphemaries certain visible differences can be noted in Table 2.2. In the case of Ragileo, this grapheme uses only one grapheme per phoneme, and on certain occasions, the sounds associated with these graphemes do not correspond to those of Spanish, so it is a little more difficult to learn than the others. The Unified has a script more similar to Castilian. Although, although most of the graphemes are the same, there are phonemes that can be considered similar, but are not the same between Castilian and Mapuzugun. Finally, the Azümchefe is a kind of intermediate point, but it also presents difficulties and differences between graphemes and phonemes in Spanish. It is used by public institutions such as CONADI.

This lack of standardization of Mapuzugun brings complications to people who are studying the language and who only master a grapheme. This also affects the processing of Mapuzugun, since there would be inconsistencies when taking data from different sources or even from the same source, especially in topics such as automatic translation or semantic analysis, where the same word could have various forms and affect learning. some model. This probably affects the current lack of basic tools in this language.

In this direction, there are currently various works related purely to Mapuzugun linguistics: descriptions \cite{zuniga:2006,smeets1989mapuche,chiguailaf1972tagmemic}, but also specific academic articles on technical aspects of the language~\cite{chiodi1999crear,olate2013interactividad,sadowsky2013mapudungun,croese2014tiempo,araya2019noun,alvarado2019nativohablantismo,sandoval2020origen} and dictionaries~\cite{catrileo2017diccionario}.

\subsection{Computational Work on Mapuzugun}
Today there are various initiatives of computational linguistics on Mapuzugun. There is an orthographic normalizer and a morphological analyzer~\cite{chandia2012dungupeyem}, but it still has some errors derived from the fact that it directly applies a series of rules without analyzing the input it receives. These are aspects that could be improved. Another aspect that could be improved is that, currently, there is no possibility of choosing the output grapheme, restricting it to only one form of writing. This is inconvenient today that there is still no agreement on the standardization of writing. This implementation was made from a set of rules through regular expressions, with a finite state transducer, which have been released on the author's website. This author is also working on a prototype morphological analyzer and spell checker, based on Xerox finite state tools. There are also corpus exploitation interfaces annotated with these same tools, created in an interuniversity master's degree in Barcelona, (coordinated by the University of Barcelona) and an automatic Mapuzugun translator is being targeted. As he is in the process of doctoral work, the results of these tools have not yet been published, but they can be reviewed in his thesis proposal.

On the other hand, there was a project called AVENUE, in which the Universidad de la Frontera, the Intercultural Bilingual Education Program and the Institute of Language Technologies of Carnegie Mellon University (CMU) collaborated. The purpose of this project was to generate simple and low-cost translators, in addition to helping to preserve the Indo-American languages. This project resulted in four products:
1. In the first place, a graphemebook for the purposes of processing and computer development of the Project.
2. A 170-hour corpus that has been transcribed, but not fully revised.
3. A translation prototype consisting of a trained example-based translator. In addition, one based on transfer rules was worked on in parallel (both with Spanish as a pair). This prototype also has a morphological analyzer. After the Avenue project, CMU also worked on the automatic improvement of translations.
 4. A spell checker that is said to contain an estimated 6,000,000 words, for OpenOffice. And that consists of two dictionaries, one for roots and the other for suffixes, which within OpenOffice's MySpell, correct a text in the typical way that the user is used to. This continues to have the limitation of the grapheme, in addition to not having the security that when writing a word in another grapheme it will convert it to the one used by the system.

In the educational field, there is software to learn Mapuzugun called MAPU from a job at the Pontificia Universidad Católica de Valparaíso that also includes voice recognition to control the application, which works correctly, but is not robust to pronunciation~\cite{troncoso2012sistema}. In this work, it also refers to another Mapuzugun-to-Spanish voice-text translation prototype, based on recordings, and to a chatbot from the Pandora project.

In addition, the CEDETI of the Pontificia Universidad Católica, which is dedicated to working on technologies for integration, has language learning software called Mapudungun mew~\cite{rosasamapudungun}.

\subsection{The three orthographies currently used}
\begin{figure*}
    \centering
    \includegraphics[scale=0.3]{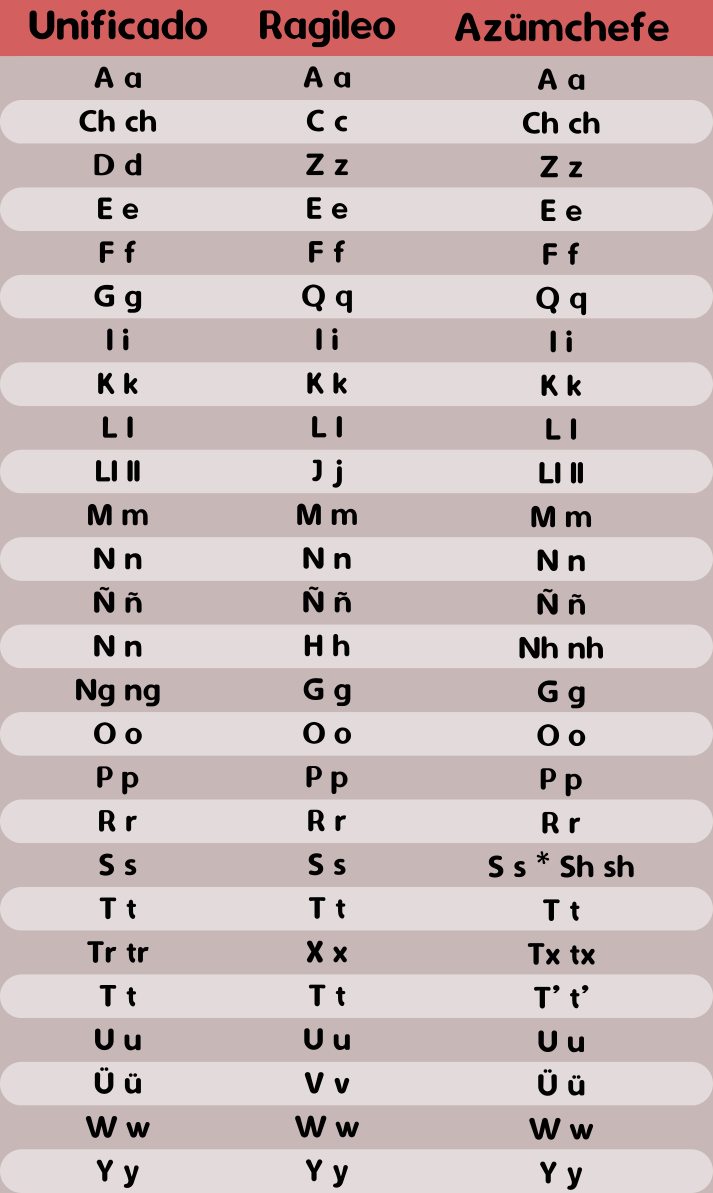}
    \caption{Comparison of the three alphabets used by the Mapuche.}
    \label{fig:alphabets}
\end{figure*}

See Figure~\ref{fig:alphabets} for a comparison.

\begin{table}[t]
    \centering
    \begin{tabular}{cc}
    \toprule
         Unificado & Ragileo \\ 
    \midrule
CH & C \\ 
D  & Z \\ 
G  & Q \\ 
L  & B \\ 
Ll  & J \\ 
N  & H \\ 
Ng  & G \\ 
Tr  & X \\ 
T  & - \\ 
Ü  & V \\
\bottomrule
    \end{tabular}
    \caption{Differences and conversion between the Unificado and Ragileo orthographies.}
    \label{tab:uniragi}
\end{table}

\begin{table}[t]
    \centering
    \begin{tabular}{cc}
    \toprule
    Unificado  & Azümchefe \\ 
    \midrule
D  & Z \\ 
G  & Q \\ 
L  & Lh \\ 
N  & Nh \\ 
Ng  & G \\ 
Tr  & Tx \\ 
T  & T’ \\ 
S  & Sh \\ 
\bottomrule
    \end{tabular}
    \caption{Differences and conversion between the Unificado and Azümchefe orthographies.}
    \label{tab:uniazu}
\end{table}

\begin{table}[t]
    \centering
    \begin{tabular}{c|c}
    \toprule
Ragileo  & Azümchefe \\ 
    \midrule
C  & Ch \\ 
B  & Lh \\ 
J  & Ll \\ 
H  & Nh \\ 
S  & Sh \\ 
X  & Tx \\ 
-  & T’ \\ 
V  & Ü \\ 
\bottomrule
    \end{tabular}
    \caption{Differences and conversion between the Ragileo and Azümchefe orthographies.}
    \label{tab:ragiazu}
\end{table}

\section{Computational Work on South American Indigenous Languages}
\label{app:sail}


\citet{mager2018challenges} review the challenges for indigenous languages in America in terms of language technologies and NLP, which is also a review of the experiences that have been had for different languages throughout the continent. Beyond Mapuzugun, it also addresses languages such as Quechua, Nahuatl, Wixarika, Shipibo Konibo, Guaraní, among others. The challenges have to do mainly with the insufficient or not well developed corpora, translations, and morphological analyzers. In addition, experiences are named in the different common tasks for NLP.


\citet{llitjos2005developing} presents the most complete process of what would be the result of the AVENUE project, whose product was a Quechua - Spanish translator. This could not be completed for the Mapuzugun case, but there is a methodology with which it could continue. 
One can also see the use of Bayesian classifiers and K nearest neighbors (k-nearest neighbors, KNN) for disambiguation in Quechua translation ~\cite{rudnick2011towards}.

Also in Quechua, the improvement of morpheme recognition from its comparison with Finnish, due to the fact that they have similar structures, especially in the agglutination part~\cite{ortega2018using}.

The closeness in typology also happens with other languages that are in Peru and the rest of the continent, such as Mexicanero, Nahuatl, Wixarika and Yorem Nokki~\cite{kann2018fortification}. Or the Mohawk and Plains Cree~\cite{arppe2016basic}, from further north.

At the University of Limerick a thesis was developed on a morphological analyzer for the Mohawk case. This is done through finite states and their training from the language data~\cite{assini2013natural}.

\citet{espichan2017language} present a study of low-resource Peruvian languages. This is done from the construction of a vector space model for languages, from bigrams and trigrams, and a matrix from "term frequency - inverse document frequency" or (TF-IDF, for its acronym, in English). It is classified by sentences and a performance of over 96\% is achieved in classification with support vector machine. Although these are good results, there is no way to know if it is exactly the orthography used or if it is just the closest.

\citet{alva2017spell} propose a corrector based on syllabification and characters for an agglutinating Peruvian language. This is done with graphs of syllables and characters from models extracted from the corpus. This method proposes three closest corrections for a misspelled word with distance metrics per character, also saving the previous corrections. This method is complete and takes into account the syllables and characters, which would be important in the case of orthographies which have subtle differences, as if they were spelling errors. Although the error can be improved (76\%), it could be a solution for the normalizer, if it is extended to multiple languages (or in this case orthographies).


\end{document}